\documentclass[11pt,a4paper]{article}
\usepackage[hyperref]{eacl2021}
\usepackage{times}
\usepackage{latexsym}
\usepackage{booktabs}
\usepackage{graphicx}
\usepackage{multirow}
\usepackage{subcaption}
\usepackage{amssymb}
\usepackage{adjustbox}

\usepackage{microtype}

\aclfinalcopy 
\def\aclpaperid{1349} 

\newcommand\blfootnote[1]{%
  \begingroup
  \renewcommand\thefootnote{}\footnote{#1}%
  \addtocounter{footnote}{-1}%
  \endgroup
}

\title{``Killing Me'' Is Not a Spoiler: Spoiler Detection Model using Graph Neural Networks with Dependency Relation-Aware Attention Mechanism}

\author{Buru Chang$^{\dagger}$ \quad Inggeol Lee$^{\ddagger}$ \quad Hyunjae Kim$^{\ddagger}$ \quad Jaewoo Kang$^{\ddagger\ast}$ \\
Hyperconnect$^{\dagger}$ \quad Korea University$^{\ddagger}$ \\
\texttt{buru@hpcnt.com} \\
\texttt{\{ingulbull,hyunjae-kim,kangj\}@korea.ac.kr}
}

\date{}

\begin{document}
\maketitle

\begin{abstract}
Several machine learning-based spoiler detection models have been proposed recently to protect users from spoilers on review websites. Although dependency relations between context words are important for detecting spoilers, current attention-based spoiler detection models are insufficient for utilizing dependency relations. To address this problem, we propose a new spoiler detection model called SDGNN that is based on syntax-aware graph neural networks. In the experiments on two real-world benchmark datasets, we show that our SDGNN outperforms the existing spoiler detection models.
\end{abstract}

\blfootnote{
\textsuperscript{$\ast$}Corresponding author.}
\blfootnote{
\textsuperscript{$\dagger$}This work was done while the author was affiliated with Korea University.}

\section{Introduction}\label{sec:introduction}

\textit{Spoilers} on review websites, which reveal critical details of the original works, can ruin an appreciation for the works.
Review websites, such as Rotten Tomato, IMDb, and Metacritic, provide self-reporting systems that tag spoiler information to warn users of spoilers.
However, since self-reporting systems depend solely on the active participation of users, they cannot handle the fast-growing volume of newly generated reviews.
During the past decade, several machine learning-based spoiler detection (SD) models have been proposed to solve the inefficiency of self-reporting systems.
\citet{guo2010finding} proposed an automatic SD model that measures the similarity between reviews and synopses of movies.
Support vector machine (SVM)-based SD models using handcrafted features have been proposed \cite{boyd2013spoiler,jeon2016spoiler}.
Recently, attention-based SD models that utilize metadata of review documents achieve state-of-the-art performance on the SD task \cite{chang2018deep,wan2019fine}.

\begin{figure}[t]
\centering
\includegraphics[width=\columnwidth]{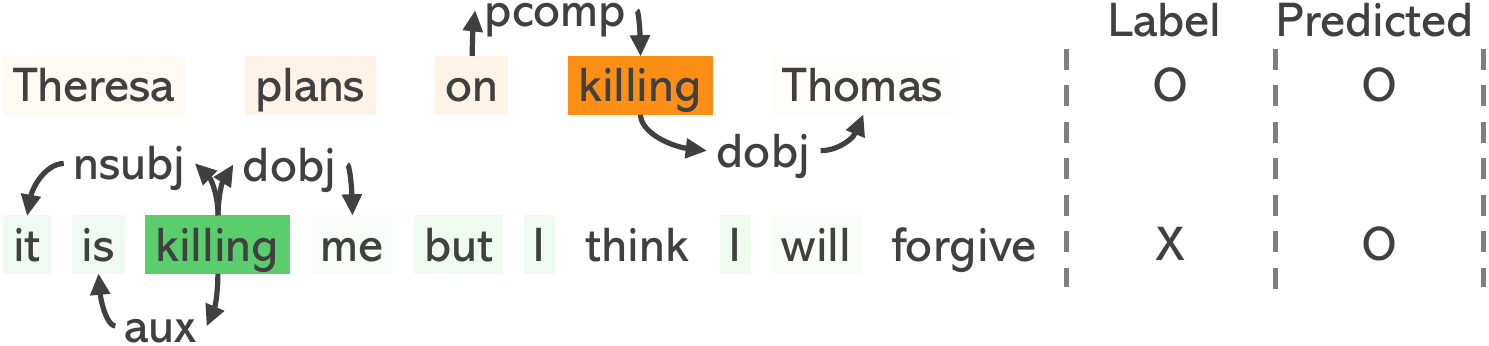}
\caption{Attention-based models focused on the word ``killing" because the word is frequently used in spoiler sentences, which results in incorrect predictions.}
\label{fig:fig_1}
\end{figure}

However, the attention-based SD models have a lack of using dependency relations between context words.
Dependency relations are useful for capturing the semantics of given sentences and detecting spoilers.
As shown in Figure \ref{fig:fig_1}, although the phrase ``killing me" is not a spoiler because the phrase is a metaphor, the attention-based SD models often focus on the word ``killing'' and classify sentences that contain the phrase ``killing me" as spoilers.
By providing the information that the word ``me" is used as the direct object of the verb ``killing," SD models can understand that the phrase is a metaphor.

In this paper, we propose SDGNN, which is a new \textbf{S}poiler \textbf{D}etection model based on syntax-aware \textbf{G}raph \textbf{N}eural \textbf{N}etworks (GNNs) \cite{marcheggiani2017encoding} for leveraging dependency relations between context words in sentences to fully capture the semantics.
With the success of GNNs 
We also propose a dependency relation-aware attention mechanism, which is a modification of the gating mechanism used by syntax-aware GNNs, to be suitable for the spoiler detection task.
In SD, considering the relative importance of dependency relations. 
However, existing syntax-aware GNN-based models compute the importance of each dependency relation individually in sentences without considering the context of the given sentence.
Our proposed dependency relation-aware attention mechanism considers the relative importance of dependency relations.
Also, we adopt a previously proposed genre-aware pooling method \cite{chang2018deep} to utilize the genre of works efficiently.
In the experiments, we demonstrate the effectiveness of SDGNN on two real-world benchmark datasets in both quantitative and qualitative ways.

\section{Our Approach}\label{sec:our_approach}
SDGNN classifies whether a given sentence $x = (w_1, w_2, \cdots, w_n)$ is a spoiler sentence.
SDGNN consists of three stages: contextualized word representation, dependency relation-aware attention mechanism, and genre-aware pooling.

\paragraph{Contextualized Word Representation}
Each word $w$ in the given sentence $x$ is represented with the pretrained word embedding vector \cite{pennington2014glove}.
We then utilize bi-directional LSTMs \cite{hochreiter1997long} to encode contextualized word representations $\mathbf{h}^{(0)} \in \mathbb{R}^{d}$.

\paragraph{Dependency Relation-Aware  Attention Mechanism}\label{subsec:dependency_relation-aware_attention}
While the gating mechanism in syntax-aware GNNs \cite{marcheggiani2017encoding,nguyen2018graph} computes the scalar weight of each dependency relation, it does not consider the relative importance of dependency relations, which varies depending on the context of the given sentence.
We present a dependency relation-aware attention mechanism that considers the relative importance of dependency relations in the given sentence.
The relation-aware attention weights are computed as follows:
\begin{equation}
\small
a^{(k)}_{L(u,v)} = g\Big(\mathbf{h}^{(k)}_u \mathbf{\overline{W}}^{(k)} {\mathbf{e}}^{(k)}_{L(u,v)} + \overline{b}^{(k)}_{L(u,v)}\Big),
\label{eq:04}
\end{equation}
\begin{equation}
\small
\hat{a}^{(k)}_{L(u,v)} = \frac{\mathrm{exp}\big(a^{(k)}_{L(u,v)}\big)}{\sum_{v'\in \mathcal{N}(u)}{\mathrm{exp}\big(a^{(k)}_{L(u,v')}\big)}},
\label{eq:05}
\end{equation}
where $\hat{a}^{k}_{L(u,v)}$ is a scalar attention weight of the dependency relation label $L(u,v)$ of the edge between word nodes $u, v$.
$g$ is the non-linear function and exp($\cdot$) is an exponential function.
$\overline{\mathbf{W}}^{(k)} \in \mathbb{R}^{d\times d}$ is the attention weight matrix for the $k$-th layer.
$\mathbf{e}_{L(u,v)}^{(k)} \in \mathbb{R}^{d}$ and $\overline{b}^{k}_{L(u,v)} \in \mathbb{R}$ are latent features of the dependency relation $L(u,v)$.

Finally, we aggregate the latent feature for each node $u$ as follows:
\begin{equation}
\small
\mathbf{h}^{(k)}_u = {f}\Big(\sum_{v\in \mathcal{N}(u)}{{\hat{a}}_{L(u,v)}^{(k)} \mathbf{W}^{(k)} \mathbf{h}^{(k-1)}_v} + \mathbf{b}^{(k)}\Big),
\label{eq:06}
\end{equation}
where $\mathbf{W}^{(k)} \in \mathbb{R}^{d\times d}$ and $b^{(k)} \in \mathbb{R}^d$ are the weight matrix and bias term, respectively, for the $k$-th layer.
$f$ is a non-linear function.
$\mathbf{h}^{(0)}$ is the outputs of the bi-directional LSTMs in the previous stage.

There are two main differences between our proposed dependency-aware attention mechanism and the gating mechanism used by syntax-aware GNNS.
First, the dependency-aware attention mechanism employs the softmax function to capture the relative importance of dependency relations, while the gating mechanism computes the scalar weights by the inner-product of latent features of words and dependency relations.
Second, the gating mechanism utilizes only three dependency relations (\textit{forward}, \textit{backward}, and \textit{self}) because of the over-parameterization issue.
On the other hand, our proposed dependency relation-aware attention mechanism utilizes all the 82 types of dependency relations without suffering from the over-parameterization issue since the weight matrix in Equation \ref{eq:06} does not depend on the number of relations.
The number of trainable parameters of SDGNN is proportionate to $d^2$ while that of syntax-aware GNNs is proportionate to $|L|\cdot d^2$, where $|L|$ is the number of relations.

\paragraph{Genre-Aware Pooling}
Genre information is useful for detecting spoilers.
To leverage genre information, we employ a genre-aware pooling method following \citet{chang2018deep}.
The genre-aware pooling computes the attention weights between the latent features of words and a genre feature captured from genre information of works.
We then obtain a latent feature vector $\mathbf{x}$ for the given sentence $x$.

\paragraph{Optimization}
We compute the spoiler probability $\hat{y}$ of the given sentence $x$ with the following the linear transformation:
\begin{equation}
\small
\hat{y} = \sigma(\textbf{w}\mathbf{x} + b),
\label{eq:10}
\end{equation}
where $\mathbf{w}$ and $b$ are trainable parameters, and $\sigma$ is a sigmoid function.
We use the weighted binary cross entropy \cite{wan2019fine} as the loss function.
\begin{equation}
\small
\mathcal{L} = -\frac{1}{|\mathcal{D}|}\sum_{x_i\in\mathcal{D}}{(y_i\mathtt{log}(\hat{y}_i)+\eta\cdot(1-y_i)\mathtt{log}(1-\hat{y}_i))},
\end{equation}
where $y$ id the ground truth of spoiler information and $\mathcal{D}$ indicates the dataset.
$\eta$ is a hyperpameter used to balance the number of spoiler and non-spoiler labels in the training data.
All the trainable parameters of SDGNN are updated by minimizing the loss function with gradient descent.

\section{Experiments}\label{sec:experiments}
\subsection{Experimental Setup}\label{subsec:experimental_setup}
\begin{table}[t]
    \centering
    \footnotesize
    \begin{tabular}{lcc}
        \toprule
        Statistics & Goodreads & TVTropes \\\midrule
        \# of Training Sentences & 14,007,593 & 11,970\\
        \# of Validation Sentences & 128,718 & 2,808\\
        \# of Test Sentences & 3,536,341 & 1,477 \\
        \# of Edge Types & 82 & 82\\
        \# of Genre & 542 & 30\\
        Avg. \# of Nodes per Sentence & 17.7 & 21.03 \\
        Avg. \# of Edges per Sentence & 33.4 & 40.06\\
        Avg. \# of Genre per Sentence & 4.95 & 2.40 \\
    \bottomrule
    \end{tabular}
    \caption{Statistics of the datasets.}
    \label{table:01}
\end{table}
\paragraph{Datasets}
We evaluated our proposed model on the two public spoiler datasets: \textbf{Goodreads} \cite{wan2019fine} and \textbf{TVTropes} \cite{boyd2013spoiler} \footnote{We obtained the datasets from Wan et al.~\shortcite{wan2019fine} and Boyd-Graber et al.~\shortcite{boyd2013spoiler}, respectively.}.
The Goodreads dataset consists of spoiler sentences on book reviews, and only 3.22\% of entire sentences are labeled as spoiler sentences.
The TVTropes dataset consists of descriptions of 884 TV programs from the TVTropes site, and 52.7\% of the descriptions are labeled as spoilers.
The statistics of the datasets are summarized in Table \ref{table:01}.

\paragraph{Baseline Models}\label{subsec:baselines}
We compared our proposed model with the following state-of-the-art SD models: \textbf{SVM} \cite{boyd2013spoiler, jeon2016spoiler}, \textbf{CNN} \cite{kim2014convolutional}, \textbf{HAN} \cite{yang2016hierarchical}, \textbf{SpoilerNet} \cite{wan2019fine} and \textbf{DNSD} \cite{chang2018deep}.
Note that the implementation details about our experiments are described in the Appendix due to space limitations.

\paragraph{Metrics}\label{subsec:metric}
We use Area Under the Receiver Operating Characteristics curve (\textbf{AUROC}) used in Wan et al.~\shortcite{wan2019fine} as an evaluation metric.
We also use an \textbf{F1} score following \citet{chang2018deep}.

\paragraph{Implemention Details}
We trained and evaluated the models on two TITAN X (Pascal) GPUs.
We implemented SDGNN using PyTorch v1.1.
We used Stanford CoreNLP \cite{manning2014stanford} to generate dependency parse trees.
We employed GloVe \cite{pennington2014glove} to represent word vectors in neural network-based models including SDGNN.
Using the validation set and grid search, we searched optimal hyper-parameters for each SD model.
All the neural network-based models were trained with the learning rate of 0.001 and the Adam optimizer \cite{kingma2014adam}.
A batch size of 1024 was used for training TextCNN, HAN, and DNSD, and a batch size of 512 was used for training SpoilerNet and SDGNN.
To prevent over-fitting, we applied L2-normalization with $\lambda$ = 1e-5 and a dropout rate of 0.5.
For TextCNN, we used 50 filters with kernel sizes of 3, 4, and 5.
For efficient training on deep learning libraries, SDGNN set the maximum number of words to 50.
For SDGNN, we used Leaky ReLU for the non-linear function $g$, and ReLU for $f$.
We set $k = 2$ for SyntacticGCN, C-GCN, and SDGNN.
We use $\eta=0.05$ for the Goodreads dataset, which is unbalanced.

\subsection{Results}\label{subsec:evaluation_results}
The experimental results are summarized in Table \ref{table:02}.
Evaluation results show that our proposed SDGNN outperforms all the baseline models including attention-based models.
This result demonstrates that our proposed dependency relation-aware attention mechanism contributes to improving SD performance.
\begin{table}[t]
    \centering
    \footnotesize
    \begin{tabular}{lcccc}
        \toprule
        \multirow{2}{*}{Models} & \multicolumn{2}{c}{Goodreads} & \multicolumn{2}{c}{TVTropes} \\
        & AUROC & F1 & AUROC & F1\\\midrule
        SVM&0.880&0.162&0.735&0.698\\
        TextCNN&0.904&0.188&0.779&0.738\\
        HAN&0.915&0.190&0.785&0.750\\
        SpoilerNet&0.924&0.194&0.808&0.768\\
        DNSD&0.928&0.199&0.818&0.788\\\midrule
        SDGNN&\textbf{0.938}&\textbf{0.210}&\textbf{0.828}&\textbf{0.801}\\\bottomrule
    \end{tabular}
    \caption{Evaluation results on two benchmark datasets. The best results are highlighted in bold.}
    \label{table:02}
\end{table}

\section{Analysis and Discussion}

\subsection{Analysis of Relative Importance}\label{subsubsec:analysis_on_syntactic_relationship-aware_attention_results}
\begin{table}[t]
    \centering
    \footnotesize
    \begin{tabular}{lcc}
        \toprule
        \multirow{2}{*}{Models} & \multicolumn{2}{c}{Goodreads}\\
        & AUROC & F1\\\midrule
        SytacticGCN&0.933&0.204\\
        C-GCN&0.923&0.193\\\midrule
        SDGNN&\textbf{0.938}&\textbf{0.210}\\\bottomrule
    \end{tabular}
    \caption{Evaluation results on the Goodreads dataset.}
    \label{table:03}
\end{table}


\begin{figure}[t]
\centering
\includegraphics[width=0.98\columnwidth]{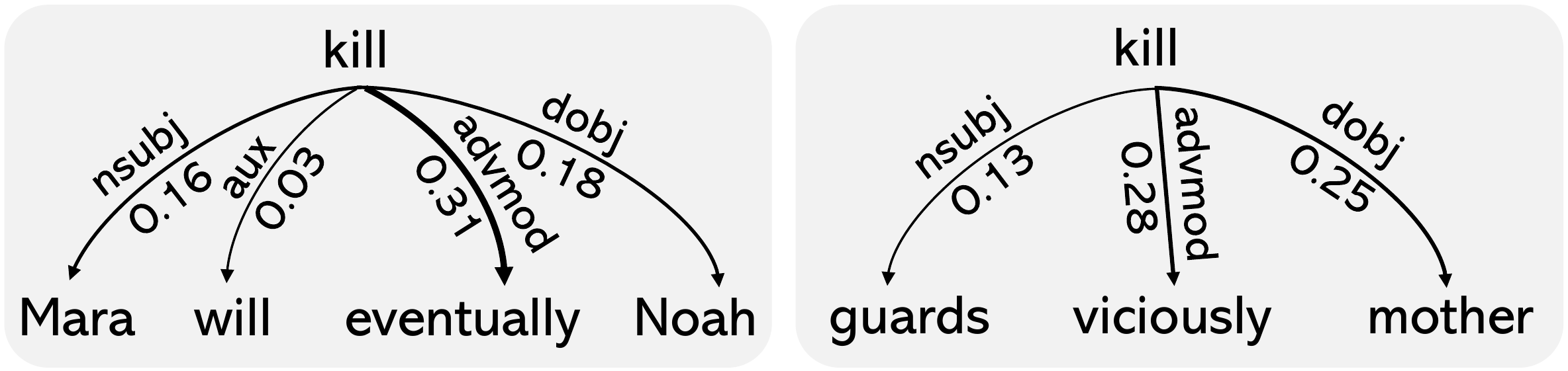}
\caption{Partial graphs of dependency parse trees with dependency relation-aware attention weights.}
\label{fig:fig_3}
\end{figure}
\begin{table*}[t]
    \centering
    \scriptsize
    \begin{tabular}{cclcc}
        \toprule
        Models & Genres & Sentences & Prediction & Label \\\midrule

        \multirow{4}{*}{DNSD}&\multirow{2}{*}{Romance}&\colorbox{orange!2}{\strut it}\colorbox{orange!2}{\strut 's}\colorbox{orange!49}{\strut killing}\colorbox{orange!7}{\strut me}\colorbox{orange!42}{\strut but}\colorbox{orange!2}{\strut i}\colorbox{orange!4}{\strut think}\colorbox{orange!2}{\strut i}\colorbox{orange!2}{\strut 'll}\colorbox{orange!68}{\strut forgive}\colorbox{orange!4}{\strut him}\colorbox{orange!3}{\strut no}\colorbox{orange!3}{\strut matter}&\multirow{2}{*}{\colorbox{red!30}{Positive}}&\multirow{2}{*}{Negative}\\
        
        &&\colorbox{orange!2}{\strut what}\colorbox{orange!2}{\strut he}\colorbox{orange!2}{\strut did}\colorbox{orange!3}{\strut or}\colorbox{orange!3}{\strut did}\colorbox{orange!3}{\strut n't}\colorbox{orange!4}{\strut do}\colorbox{orange!2}{\strut .}&&\\\cmidrule(lr){2-5}
        &Fantasy&\colorbox{orange!3}{\strut the}\colorbox{orange!50}{\strut villains}\colorbox{orange!6}{\strut are}\colorbox{orange!10}{\strut decidedly}\colorbox{orange!57}{\strut vicious}\colorbox{orange!7}{\strut and}\colorbox{orange!4}{\strut in}\colorbox{orange!4}{\strut some}\colorbox{orange!7}{\strut cases}\colorbox{orange!56}{\strut insane}\colorbox{orange!3}{\strut .}&\colorbox{red!30}{Positive}&Negative\\

\midrule
        
        \multirow{4}{*}{SDGNN}&\multirow{2}{*}{Romance}&\colorbox{orange!3}{\strut it}\colorbox{orange!2}{\strut 's}\colorbox{orange!57}{\strut killing}\colorbox{orange!66}{\strut me}\colorbox{orange!10}{\strut but}\colorbox{orange!3}{\strut i}\colorbox{orange!3}{\strut think}\colorbox{orange!3}{\strut i}\colorbox{orange!4}{\strut 'll}\colorbox{orange!16}{\strut forgive}\colorbox{orange!3}{\strut him}\colorbox{orange!4}{\strut no}\colorbox{orange!3}{\strut matter}&\multirow{2}{*}{\colorbox{green!30}{Negative}}&\multirow{2}{*}{Negative}\\
        
        &&\colorbox{orange!2}{\strut what}\colorbox{orange!3}{\strut he}\colorbox{orange!3}{\strut did}\colorbox{orange!4}{\strut or}\colorbox{orange!3}{\strut did}\colorbox{orange!3}{\strut n't}\colorbox{orange!2}{\strut do}\colorbox{orange!2}{\strut .}&&\\\cmidrule(lr){2-5}
        
        &Fantasy&\colorbox{orange!2}{\strut the}\colorbox{orange!56}{\strut villains}\colorbox{orange!68}{\strut are}\colorbox{orange!4}{\strut decidedly}\colorbox{orange!15}{\strut vicious}\colorbox{orange!6}{\strut and}\colorbox{orange!3}{\strut in}\colorbox{orange!4}{\strut some}\colorbox{orange!3}{\strut cases}\colorbox{orange!11}{\strut insane}\colorbox{orange!2}{\strut .}&\colorbox{green!30}{Negative}&Negative\\
\midrule
        
    \end{tabular}
    \caption{Visualization of attention scores from DNSD, SpoilerNet, and SDGNN on test data.}
    \label{table:04}
    \vspace{-5mm}
\end{table*}


To further demonstrate the usefulness of the relative importance of dependency relations, we conducted quantitative and qualitative analysis.
\paragraph{Quantitative} We compared SDGNN with the more syntax-aware GNN-based models, SyntacticGCN \cite{marcheggiani2017encoding} and C-GCN \cite{zhang2018graph}.
We trained and evaluated the models on the Goodreads dataset.
We utilize contextualized word representations and the genre-aware pooling method to SyntacticGCN and C-GCN.
The evaluation results are summarized in Table \ref{table:03}.
Our proposed SDGNN outperformed SyntacticGCN and C-GCN.
This result demonstrates that our proposed attention mechanism is effective by considering the relative importance of dependency relations.
Although SDGNN significantly reduced the number of parameters, SDGNN achieved better results compared to SyntacticGCN and C-GCN.

\paragraph{Qualitative} In Figure \ref{fig:fig_3}, the attention weights of the adverbial modifier (advmod) linked to the words ``eventually" and ``viciously" are high, which indicates that adverbial modifiers frequently can be important hints for detecting spoilers.
In the right partial graph, the attention weight of the (dobj) is relatively higher than that in the left partial graph.
Since the word ``mother" is not typically used as the object of the word ``kill" in the original works, the phrase ``kill mother" is a critical hint in detecting spoilers, and SDGNN effectively captures the phrase.


\subsection{Case Study}\label{subsec:case_study}
We sample several sentences from the test set of the Goodreads dataset to explore how the models detect spoilers.
Table \ref{table:04} shows the visualization of attention scores in the pooling layer obtained by DNSD and SDGNN, respectively.
The first sentence contains the verb ``killing," but it is not a spoiler sentence because the phrase ``killing me" is a metaphor.
In this case, DNSD failed to correctly classify the sentence since DNSD cannot fully capture the semantics of the sentence.
On the other hand, SDGNN focused on not only the word ``killing" but also on the word ``me" and classified the sentence correctly since SDGNN employs the dependency relation (dobj) between the word ``me" and the word ``killing".

The second sentence is a non-spoiler because it is obvious that villains are vicious in most original works.
DNSD classified the sentence as a spoiler because the model solely focused on individual words such as ``villains", ``vicious", and ``insane", rather than the understanding of the overall semantics of the sentence.
On the other hand, SDGNN classified the sentence correctly as the word ``are" is used to describe characters in many cases, and SDGNN understands the semantics of the sentence.

\subsection{Discussion}
\paragraph{Dependency Parsing on User-Generated Texts}
The spoiler datasets are user-generated texts, which are intrinsically noisy.
To examine the influence of noises on dependency parsing results and the performance of SDGNN, we sampled 100 sentences from Goodreads. 
We manually classified whether the sentences are noise or not, and 28 of 100 sentences were classified as noisy sentences. 
Dependency parsing results on well-structured sentences seem good, but dependency parsing results on noisy sentences are poor.
However, there is no significant gap in performance. 
SDGNN achieved 85.7\% accuracy on noisy sentences and 87.5\% accuracy on well-structured sentences. 
Since our proposed dependency relation-aware attention mechanism of SDGNN filters noisy information, SDGNN could detect spoilers even on noisy sentences.

\paragraph{Subjectivity in Judging Spoilers}
Since judging a sentence as a spoiler is a subjective task, label inconsistency occurs in spoiler datasets crawled from self-reporting systems.
\citet{guo2010finding} found that 23\% of the labels of their manually labeled data is different from the original labels of IMDb reviews.
One of the ways to mitigate label inconsistency is to solidify the definition of a spoiler.
Although the TV Tropes site defines spoilers, efforts should be made for a more rigorous and linguistic definition in future studies.
Another possible way is to employ reviewers' information in detecting spoilers.
\textit{Reviewer biases} of SpoilerNet can alleviate label inconsistency between users.

\section{Conclusion}\label{sec:conclusion}
In this paper, we proposed a novel spoiler detection model called SDGNN which is based on syntax-aware GNNs that utilize dependency relations between context words.
We also proposed a dependency relation-aware attention mechanism for considering the relative importance of dependency relations.
In the experiments, our proposed SDGNN model achieved the state-of-the-art performance on two spoiler datasets.
Our experimental results demonstrate the effectiveness of dependency relations in the spoiler detection task and our dependency relation-aware attention mechanism.

\section*{Acknowledgments}

This research was supported by National Research
Foundation of Korea (NRF-2020R1A2C3010638).
This research was also supported by the MSIT (Ministry of Science and ICT), Korea, under the ICT Creative Consilience program (IITP-2020-0-01819) supervised by the IITP (Institute for Information \& communications Technology Planning \& Evaluation).

\bibliography{anthology, eacl2021}
\bibliographystyle{acl_natbib}

\appendix

\end{document}